\documentclass[letterpaper, 10 pt, conference]{ieeeconf}  

\IEEEoverridecommandlockouts                              

\usepackage{hyperref}
\usepackage{graphics} 
\usepackage{epsfig} 
\usepackage{mathptmx} 
\usepackage{times} 
\usepackage{amsmath} 
\usepackage{amssymb}  
\usepackage{subfigure}
\usepackage{tabularx,booktabs}
\usepackage{multirow}
\usepackage{multicol}

\title{\LARGE \bf
Scene Graph for Active Exploration in Cluttered Scenario
}

\author{Yuhong Deng$^{1,\dagger}$, Qie Sima$^{2,\dagger}$, Huaping Liu$^{2,*}$ and Fuchun Sun$^{2}$
\thanks{$^\dagger$ indicates the authors with equal contributions.}
\thanks{ $^{1}$ National University of Singapore, Singapore}
\thanks{ $^{2}$ Department of Computer Science and Technology, Tsinghua University, Beijing, China}
\thanks{$^*$Corresponding author}%
}

\begin{document}

\maketitle
\thispagestyle{empty}
\pagestyle{empty}

\begin{abstract}

Robotic question answering is a representative human-robot interaction task, where the robot must respond to human questions. Among these robotic question answering tasks, Manipulation Question Answering (MQA) requires an embodied robot to have the active exploration ability and semantic understanding ability of vision and language. MQA tasks are typically confronted with two main challenges: the semantic understanding of clutter scenes and manipulation planning based on semantics hidden in vision and language. To address the above challenges, we first introduce a dynamic scene graph to represent the spatial relationship between objects in cluttered scenarios. Then, we propose a GRU-based structure to tackle the sequence-to-sequence task of manipulation planning. At each timestep, the scene graph will be updated after the robot actively explore the scenario. After thoroughly exploring the scenario, the robot can output a correct answer by searching the final scene graph. Extensive experiments have been conducted on tasks with different interaction requirements to demonstrate that our proposed framework is effective for MQA tasks. Experiments results also show that our dynamic scene graph represents semantics in clutter effectively and GRU-based structure performs well in the manipulation planning.

\end{abstract}

\section{INTRODUCTION}
\label{section: intro}

People have long anticipated that one day embodied robot can directly receive human questions based on natural language and actively interact with the real environment to respond and give an answer~\cite{najafi2018robot}, which reflects intelligent interactions between robot, human, and environment~\cite{prati2021use}. Recently, active manipulation have been widely used in embodied robot tasks to enable the agent to retrieve more information from environment. Manipulation question answering (MQA) is a new proposed human-robot interaction task where the robot must perform manipulation actions to actively explore the environment to answer a given question\cite{deng2020mqa}. However, proposed methods for MQA task only focus on one specific task related to counting questions\cite{uc2022survey,li2021robotic}. Considering that the form of human-computer interaction should be varied, we extend the variety of question types in this work and aim to design a general framework for MQA tasks. \par
Different from previous robotic question answering tasks~\cite{antol2015vqa,das2018embodied,gordon2017iqa}, manipulation question answering poses two new challenges. The first challenge comes from the semantic understanding of clutter. The application scene of MQA is cluttered with many unstructured layouts of various objects. Thus, it is not easy to understand the semantic information in the scene, which is crucial for the robot to give a correct answer. The second challenge lies in the planning of manipulation based on semantics; the mapping from vision and language to manipulation sequences is not direct. How to obtain manipulation policies that can help the robot actively explore the scene effectively for question answering is still an open challenge.\par

\begin{figure}[t]
	\centering
	\includegraphics[width=\linewidth]{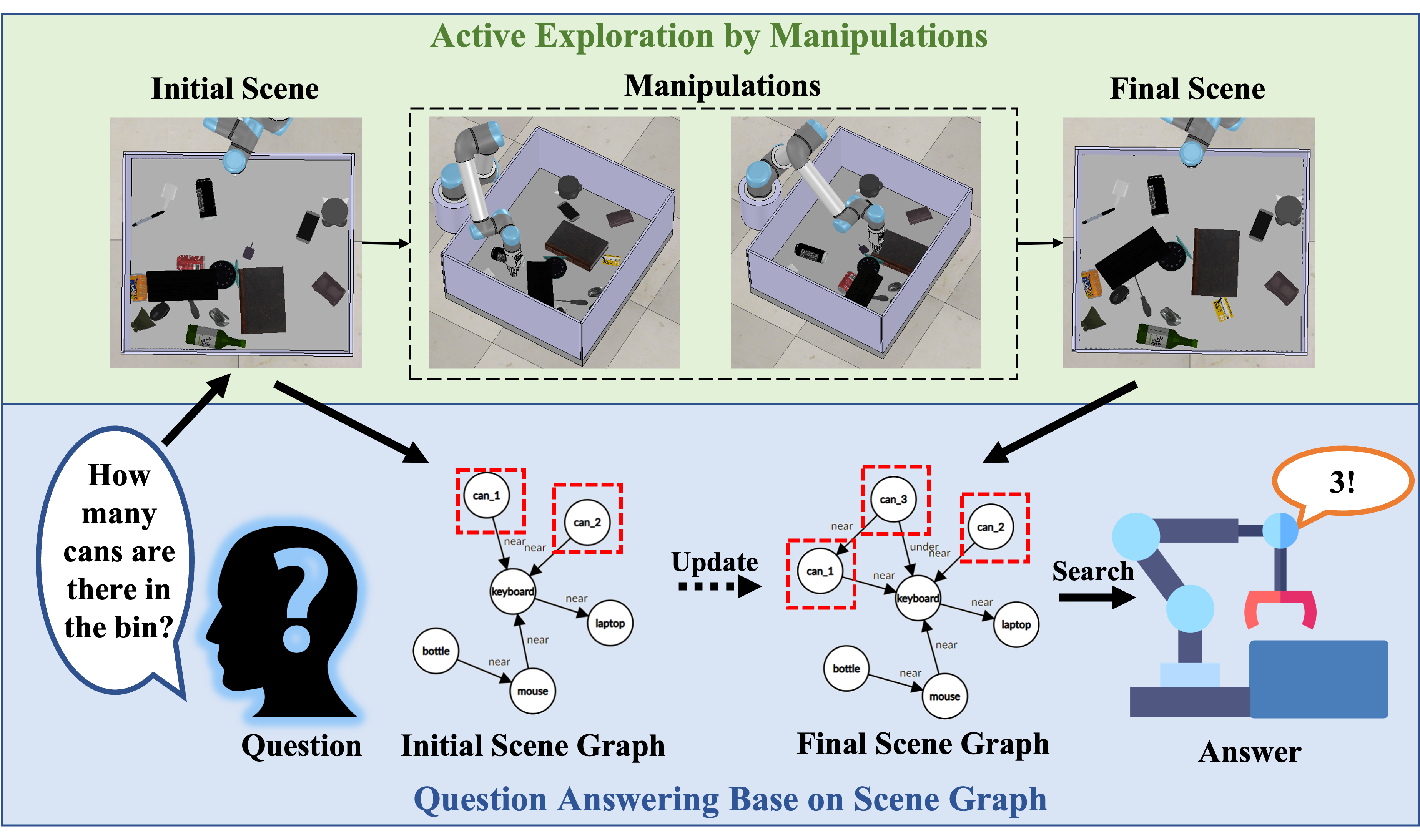}
	\caption{After receiving the question, the robot will understand the semantic information hidden in clutter by the initial scene graph. A manipulation sequence will be used to active explore the clutter and obtain the final scene graph contains sufficient information for answering the given question. And the robot can output the answer by searching the final graph. }
	\label{fig:summary}
\end{figure}

To address semantic understanding challenges, an efficient representation method for cluttered scenes, delivering accurate and ample semantic data, is essential. In recent years, scene graphs with semantic labels have gained traction in robot manipulation tasks, offering a means to convey semantic insights within chaotic scenarios~\cite{kim20193,das2021semantic,kenfack2020robotvqa}. Objects are viewed as nodes, and the spatial object relationships are represented as graph edges. Hence, we propose a dynamic scene graph to unveil obscured semantics within cluttered scenes. To construct this graph, we employ a Mask-RCNN detector for object node recognition. Given the prevalent relationships in MQA scenes, specifically, overlapping and proximity, we introduce two types of edges: 'above/below' and 'nearby' in the scene graph.  The intricate arrangement of objects in clutter often results in significant stacking and overlap, elevating semantic understanding complexity. Passive perception alone falls short in acquiring precise scene graphs; our robot also possesses active perception capabilities. As depicted in Figure \ref{fig:summary}, the robot actively explores the cluttered scene through object manipulation, enabling continuous scene graph updates with additional semantic details concealed within clutter. After a comprehensive exploration of the cluttered scene, the final scene graph captures the established semantic information accurately. Subsequently, the robot can answer queries by querying the final graph.\par
For manipulation planning tasks, learning high-level manipulation related to semantics in vision and language without labeled data is challenging. In our simulation-based training, the oracle agent extracts information about target objects from the given question, including their positions and overlapping conditions. This oracle agent generates manipulation policies that offer crucial semantic insights for answering questions. Consequently, we cast the manipulation planning task as a sequence-to-sequence problem and train the planning model through imitation learning. Our manipulation planning model is an end-to-end GRU-based architecture that takes vision and language information as input and produces the manipulation sequence as output.\par
We have conducted extensive experiments to analyze the performance of our proposed solution framework based on the scene graph and GRU-based structure. Our framework is general and effective for multiple MQA tasks. Furthermore, scene graphs and GRU-based structures can improve the performance in MQA tasks compared to baseline models. The contributions of this paper can be summarized as follows:
\begin{itemize}
    \item We introduce the dynamic scene graph into the manipulation task in cluttered scenarios as a solution to record the spatial information of the time-variation clutter environment.
	\item We provide an end-to-end framework to tackle semantic manipulation tasks in cluttered scenarios which utilize an imitation learning method for embodied exploration and a dynamic scene graph QA module for semantic comprehension.
	\item We conduct experiments in the MQA benchmark to validate our proposed framework's effectiveness in a cluttered environment.
\end{itemize}


This paper is organized as follows. The related works about scene graphs and manipulation tasks in the cluttered environment are investigated in Section \ref{section:related}. The proposed solution framework is presented in Section \ref{section: model}. Section \ref{section: experiment} is about experiments. Finally, we make a conclusion of the paper in Section \ref{section: conclusion}.

\section{RELATED WORK}
\label{section:related}
\subsection{Manipulation Task in Clutter Scenarios}
Various task forms have arisen to evaluate autonomous agents' interaction capabilities within cluttered environments. Grasping, with potential applications in industrial settings, has been the most extensively studied manipulation task in cluttered scenarios. This research line primarily concentrates on detecting optimal grasping poses in cluttered scenes~\cite{schwarz2018rgb,zhang2019roi,kiatos2019robust,deng2019deep}. Another challenging task in cluttered surroundings is object layout rearrangement. Successfully accomplishing this task necessitates the robot to not only accurately detect and localize every object but also comprehend their spatial relationships~\cite{cheong2020relocate,batra2020rearrangement}. However, both grasping and rearrangement tasks are devoid of cognitive objectives.\par

In recent years, researchers have directed increased attention to robotic manipulation tasks associated with linguistic interaction. In these tasks, manipulation serves to enhance the robot's cognitive abilities and facilitate linguistic interaction. Zhang et al. integrate visual grounding with grasping, framing it as a sequential task wherein the manipulator must precisely locate and pick up an object from a set of objects of the same category by posing human-generated questions~\cite{zhang2019roi}. Zheng et al. introduce a compositional benchmark framework for Visual-Language Manipulation (VLM), requiring the robot to perform a series of prescribed manipulations based on human language instructions and egocentric vision~\cite{zheng2022vlmbench}. Deng et al. present the Manipulation Question Answering (MQA) task, wherein the robot must find answers to posed questions by actively engaging with the environment through manipulation. Given MQA tasks' high demands on robotic manipulation and perception capabilities, there is currently no comprehensive MQA task framework available~\cite{deng2020mqa}.\par

\subsection{Semantic Understanding in Robotic Manipulation}
Recently, robot manipulation tasks put forward higher requirements for the robot's perception and semantic understanding abilities of the environment~\cite{li2017reinforcement}.  Some works explore the utilization of more information, including semantics and spatial relationship of objects in clutter. Therefore,  manipulation tasks in the form of linguistic interaction have been raised in recent years~\cite{das2017visual,kenfack2020robotvqa,zhang2021invigorate}. However, most of the cluttered scenes proposed in this work does not have enough objects with diversified shape, size, color, and other attributes. In scenes of MQA tasks, the diversified objects and the overlapping caused by the layout of objects increase the challenging level for perception and manipulation.\par
 To record the unstructured layout of objects in cluttered scenes, scene graphs with semantic labels have been widely used as a state representation in robot manipulation tasks in past decade~\cite{xu2017scene,rosinol20203d,sieb2020graph}. Kim et al. introduce a 3D scene graph as an environment model to represent environment where robot conduct manipulation tasks~\cite{kim20193}. Das et al. utilize the semantic scene graph as a method to explain the failure in robot manipulations~\cite{das2021semantic}. Meanwhile, other works take scene graphs as intermediate results for high-level comprehension or task planning. Kumar et al. introduce a learning framework with GNN-based scene generation to teach a robotic agent to interactively explore cluttered scenes~\cite{kumar2021graph}. Kenfack et al. propose RobotVQA, which can generate the scene graph to construct a complete and structured description of the cluttered scene for the manipulation task~\cite{kenfack2020robotvqa}. Zhu et al. implement a two-level scene graph representation to train a GNN for motion planning of long-horizon manipulation task~\cite{zhu2021hierarchical}. We introduce the scene graph to record and comprehend the time-variant cluttered scenarios to tackle the semantic understanding problem.

\begin{figure*}[ht]
	\centering
	\includegraphics[width=0.9\linewidth]{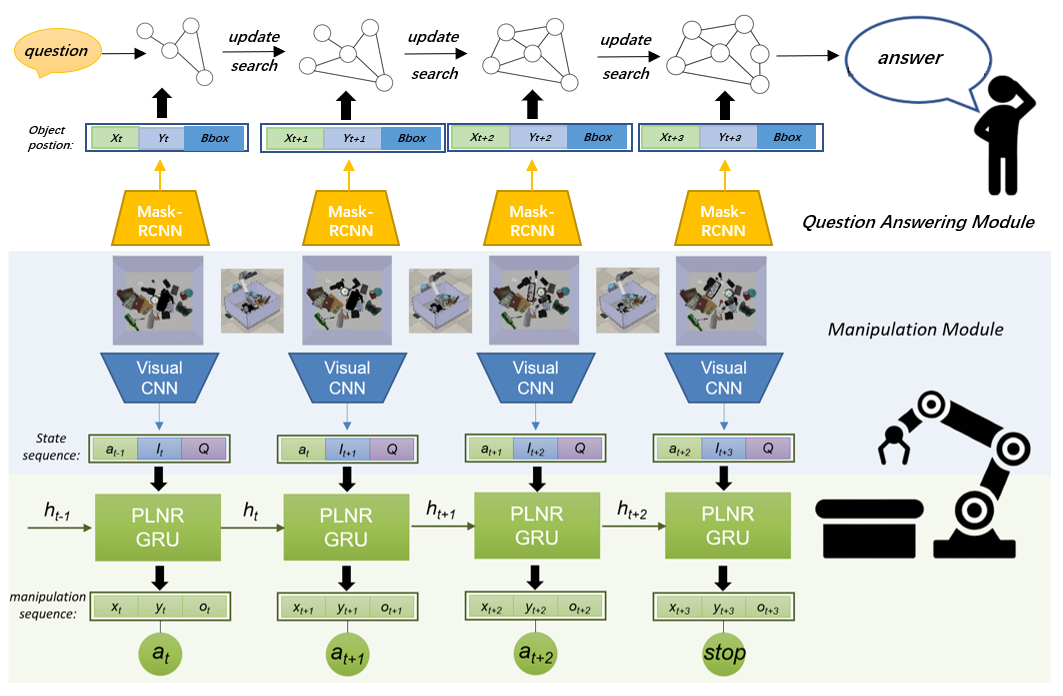}
	\caption{The architecture of the proposed MQA model. The upper part demonstrates our QA part based on the dynamic scene graph and the bottom part demonstrates our Manipulation part based on a GRU-backboned imitation learning model}
	\label{fig:action}
\end{figure*}

\section{PROPOSED MODEL}
\label{section: model}

As the MQA task is a multi-modal problem, our system consists of two parts: the manipulation module and the QA module. When a new MQA task starts, the manipulation module will start first. The manipulation module will use the RGB-D image of the scene and the question to generate a set of manipulations to explore the environment. Meanwhile, the scenario comprehension module passes the scene frame to the downstream QA module after every manipulation step until the question can be answered (the manipulation module decides when to stop). Then, the QA module will answer based on the above RGB frames and the question. (Fig.\ref{fig:action}) 

\subsection{Manipulation Module}
The task of the manipulation module is a sequence-to-sequence task. The manipulation module should output a manipulation sequence to explore the scenes based on the scene state sequence. The state sequence is updated by robotic manipulation, so the manipulation at time t can only be predicted based on the states before time t. The state at the time t comprises the current visual feature and the question. In addition, we add the last manipulation to the current state to improve the prediction performance. These feature sequences are encoded into a vector sequence. Then, we use an encoder-decoder structure (GRU encoder and linear decoder) to complete the sequence-to-sequence task.
\subsubsection{State Sequence Encoding}
Inspired by~\cite{das2018embodied}, we also encode the RGB image $I$ obtained from the Kinect camera with a CNN network $g(\cdot)$, which produces an embedded vector $v= g(I)$. The CNN model had been pre-trained under a multi-task pixel-to-pixel prediction framework. The natural language questions are encoded with 2-layer LSTMs. \par
\subsubsection{manipulation sequence encoding}
In the MQA task, the manipulation is pushing. So the manipulation should be a pushing vector which consists of pushing position $(x,y)$, pushing directions $o$, and pushing distance $d$. The action space is too complex. In order to ensure the feasibility of model learning, We deal with the action space as follows:
\begin{itemize}
\item Fixing the distance and choose a direction from eight fixed directions, the robot can push the object from 8 directions with a fixed distance or stop the task. 
\item after the first step, the size of the action space is $H*W*8$, where $H$ and $W$ represents the height and width of the RGB image. In order to further simplify action space, we sampled the action space every 8 pixels. Then the size of the action space is $\frac{1}{8}H*\frac{1}{8}W*8$.
\item Finally, we decouple the coupled action space to $x$,$y$,$o$, reduce the space size to $\frac{1}{8}H+\frac{1}{8}W+8$. \par
\end{itemize}

\subsubsection{Sequence to Sequence Model}
We use an encoder-decoder structure based on GRU to complete the sequence-to-sequence task. Compared with other models based on recurrent layers (RNN and LSTM) and the model based on attention mechanisms (transformer~\cite{attention}), our model performs better.
The transformer has recently performed well in many natural language processing sequences to sequence tasks. Our algorithm performs better than the transformer, showing our model design's rationality and effectiveness. The result of the comparative experiment is shown in \ref{section:ablation study}.

\subsubsection{Imitation Learning}
Since the ground truth answers to the questions in the simulation environment are provided, we can obtain the best action behavior for the robot to explore the scene. Therefore, we adopt an imitation learning methodology to train the manipulator to mimic the best action behavior. It is essential to define a metric to evaluate the generated manipulations. In this work, we design an imitation policy based on the least action step metric.\par

In the least action steps metric, the least steps actions are considered the best. For the EXISTENCE question, no action will be taken if the query object does not exist. Otherwise, conditioned on whether the query object can be seen directly, the robot will take no action or remove one occluding object by pushing directly. The procedure is similar to the EXISTENCE question for the COUNTING question, and the robot will remove all occluding objects. For the SPATIAL question, the robot will take no action if there is no spatial relationship between two query objects. Otherwise, the robot will push the query object on top away in order to find the answer. \par

\subsubsection{Training Details}
We have decoupled the action space, which is divided into three parts: $x$, $y$, and $o$. The total loss $loss$ is the sum of the classification loss of these three parts:
\begin{equation}
    loss = 0.25*loss_{x}+0.25*loss_{y}+0.5*loss_{o}
\end{equation}
where $loss_{x}$, $loss_{y}$, $loss_{o}$ are the classification loss of $x$, $y$, $o$, respectively. We set the weights of the three parts to 0.25, 0.25, and 0.5 because the pushing direction has more influence than the pushing position in our task.\par

\subsection{Question Answering Module}
We build a VQA model based on a dynamic scene graph of the bin scenario.  By introducing the scene graph, we can abstract the cluttered bin scenario into a graph structure and record the manipulations by updating it.  The QA module will be executed when the manipulation module generates $stop$ action or achieve max steps.  Our QA model architecture is shown in the upper of figure \ref{fig:action}.  In every task, a sequence of scene frames and the corresponding question is input.  For each sample, we generate a scene graph based on the first frame and priors about attributes of objects.  Thus, we can abstract the manipulations between frames into updating the scene graph.  
\begin{figure}[t]
	\centering
	\includegraphics[width=0.95\linewidth]{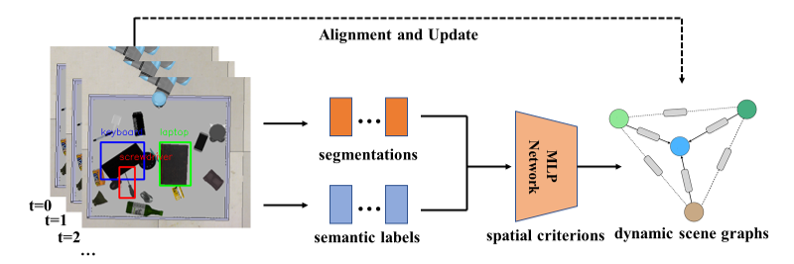}
	\caption{The architecture of the scene graph generation model. The information in segmented frame of every time step is forward processed into nodes and edges of scene graphs.}
	\label{fig:scene_generation}
\end{figure}
\subsubsection{Scene Graph Generation}
For scene graph generation, we resort to the semantic segmentation of input manipulation scenes to build the scene graph.
To get the semantic segmentation of frames, we implement an object detector based on a Mask-RCNN model with ResNet-FPN backbone to label objects in input scenes.  The Mask-RCNN model is trained on a selected subset of the COCO dataset with specific categories of objects visual in our cluttered scenes.  After rendering the segmented frames, the model generates the scene graph by adding detected objects as nodes into the graph structure and connecting nodes with edges representing the spatial relationship between different objects. The architecture of the scene graph generation model is demonstrated in Fig. \ref{fig:scene_generation}.

\begin{figure*}[ht]
        \vspace{0.2cm}
	\centering	\includegraphics[width=0.85\linewidth]{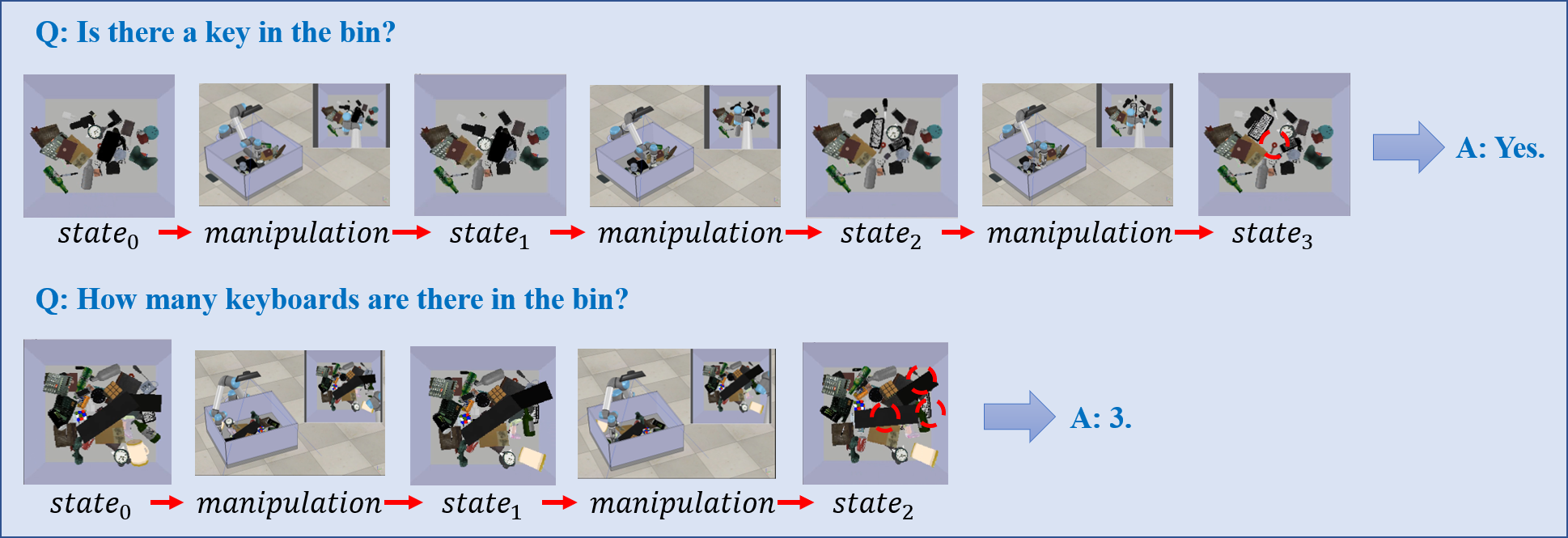}
	\caption{The performance of our MQA system: When the robot receives a question, it will take a series of manipulations to better understanding the scene. The robot will output an answer when the exploration is enough for question-answering.}
	\label{fig:ex_resu}
\end{figure*}

We decide the spatial relation between every object pair with geometric parameters of bounding boxes masked by the upstream Mask-RCNN detector.  Considering the overlapping and stacking in our scenes,  we only introduce two kinds of spatial relations: above/below and nearby. 
During the updating process of the scene graph, we propose two indicators to decide the spatial relation between two objects: overlap rate $IoU$ and normalized distance $l$.  For each object pair, we calculate these two indicators as below:
\begin{equation}
    IoU = \dfrac{S_{overlap}}{S_{union}}
\end{equation}
$S_1$,$S_2$ represent the area of bounding boxes and $S_{overlap},S_{union}$ represents the overlapping and total union area of two bounding boxes.
\begin{equation}
    l = \dfrac{d_{center}}{max(L_{1},L_{2})}
\end{equation}
$L_1$,$L_2$ represent the length of diagonals of bounding boxes and $d_{center}$ represents the Euclidean distance between geometric centers of bounding boxes.
We decide the spatial relation between two objects with following criterion: 
\begin{itemize}
  \item [(1)] $IoU\geq0.5$ which means two objects overlapping with each other in large area credits to relation \textbf{'above/below'}. 
  \item[(2)] $IoU<0.5$ while $l<0.5$ which means two objects don't overlap much but are close enough credits to relation \textbf{'above/below'}. For instance, a pen leans down at the edge of a notebook.
  \item[(3)] $IoU<0.5, 0.5\leq l <1$ which means two objects almost don't overlap but are located in surroundings of each other credits to relation \textbf{'nearby'}. 
  \item[(4)]  $IoU<0.5,  l \geq 1$ which means two objects are far away from each other credits to relation \textbf{'None'}.
\end{itemize}
The above spatial criterion is encoded in a two layer MLP network to process the input segmentations and semantic labels into spatial relationships which will be added into scene graphs as edges.
\subsubsection{Answer Prediction}
 After processing every frame, we choose the node with the most connected edges which represents the object with most surrounding objects as the key node. As shown in Fig.\ref{fig:scene_generation}, we align the key node to the corresponding node in the scene graph representation of frame at last time step.  By the alignment of scene graphs at every time step, we record the relocation of objects caused by manipulations in abstract representations which we note dynamic scene graph. \par
 The questions are encoded as word embedding vectors and classified into 3 types. Then the answer generator will retrieve the related information in scene graph. For COUNTING and SPATIAL questions, we enumerate the node and egde list to give out the number of the proposed object in the scene graph.  For the EXISTENCE questions, we take the proposed object as the key node and do Breadth first search (BFS) around it in the scene graph to examine the existence of proposed objects or pairs.

\section{EXPERIMENTS}
\label{section: experiment}
\subsection{Evaluation Experiment of Proposed MQA System }
\label{section:evaluation}
We conduct the evaluation experiment of our QA model separately in easy and hard test sets and the results are presented below.
\begin{table}[ht]
	\centering
	\renewcommand\arraystretch{1.2}
	\fontsize{7}{8}\selectfont
	\caption{Evaluation of QA model with different performance measures}
	\begin{tabular}{ccccccccc}
		\toprule
		\multicolumn{1}{c}{ Measures}&
		\multicolumn{3}{c}{Easy Scenes}&
		\multicolumn{3}{c}{Hard Scenes}\cr
    	\cmidrule(lr){2-4} \cmidrule(lr){5-7}
		&EXIST&COUNT&SPATIAL&EXIST&COUNT&SPATIAL\cr
		\midrule
		Precision &0.591&0.982&0.609&0.502&0.927&0.650\cr
	    Recall &0.553&0.745 &0.622&0.486&0.656 &0.433\cr
		Accuracy &0.863&0.945&0.611&0.834&0.799&0.400\cr
		\bottomrule
	\end{tabular}
	\label{table:test_results}
\end{table}

The above results show the efficacy of our proposed method for QA problems. Specifically, the model exhibits superior performance in EXISTENCE questions, particularly in terms of precision and recall. Conversely, the model's performance metrics are notably lower for COUNTING and SPATIAL questions. We attribute this phenomenon to two primary factors. Firstly, in our scenes, especially the challenging ones, most object types are present in the input scenes, potentially causing a model bias towards predicting the proposed object type exists. Secondly, the manipulator's actions tend to be more 'slash'-oriented than 'pick'-oriented, leading to the relocation of multiple objects simultaneously which is hard to accurately generating the scene graph and aligning the key node correctly between frames.\par

To validate the effect of manipulations, we restrict the max manipulation steps to examine whether the trained model can give a correct answer if there are not enough operations. We choose max steps 0 (no manipulation), 1, and 5 (default max steps) to conduct our validations. The results are presented in table \ref{table:testing_result}.
\begin{table}[h]
	\centering
	\fontsize{7}{8}\selectfont
	\caption{Prediction Accuracy of QA model with different max steps}
	\begin{tabular}{p{2.5cm}<{\centering}ccc}
		\toprule
		\multicolumn{1}{c}{Max Step}&
		\multicolumn{1}{c}{Easy Scene}&
		\multicolumn{1}{c}{Hard Scene}\cr
		\midrule
		0 &0.685&0.543 \cr
		1&0.804&0.785\cr
		5&0.875&0.821\cr
		
		\bottomrule
	\end{tabular}
	\label{table:testing_result}
\end{table}
The results show a significant drop in accuracy when the max steps get less. Thus, the necessity of manipulations to get correct answers can be validated.

Two examples of the evaluation experiment are shown in Fig.\ref{fig:ex_resu}. The first task is the EXISTENCE question of the key. There is no key that can be seen at first. The robot finds a suspected key after pushing the clock away. Then the robot clears the area around the suspected key to make more reliable judgments. Finally, the robot gives the "Yes" answer. The second task is a COUNTING task of keyboards. There are two visible keyboards at first. The robot clears the area around the third keyboard because only a small part of the third keyboard is visible. After the third keyboard is almost visible and there is no suspicious area that may hide a keyboard, the robot gives the answer 3.



\subsection{Ablation study}
\label{section:ablation study}
In order to verify the performance of different sequence to sequence models in our manipulation generated task, we compared the imitation performance of four kinds of models (Transformer encoder, RNN encoder, LSTM encoder and GRU encoder).
\subsubsection{Experiment metric}
Considering that the position and direction of pushing are independent, we will evaluate them separately. The position error is defined as the distance between the center point of the pushing action vector output by the model and the center point of the imitated action vector. And we normalized this distance.
\begin{equation}
    dis_e = \dfrac{\sqrt{(x_o-x_i)^2+(y_o-y_i)^2}}{224}
\end{equation}
where $(x_o,y_o)$ is the center point of  push action vector output by the model, and
$(x_o,y_o)$ is the center point of the imitated action vector. 224 is the pixel width and height of the images. The direction error is defined as the angle between above 2 vectors. And we normalized it by dividing it by 180 degree.

\subsubsection{Experiment result}
We then evaluated imitation learning on ten new scenes, divided into five easy and five hard sets. The results are presented in TABLE \ref{table:action_imitation}. Notably, among the four seq2seq models, the transformer encoder exhibits the weakest performance, despite its higher computational parallelization. Our manipulation task primarily relies on the current and preceding states, favoring sequential computation in recurrent layers over the parallel nature of transformers. Unlike the NLP domain, our robot exploration task emphasizes temporal sequences. Among the three models based on recurrent layers, GRU effectively utilizes state sequence information preceding the current time, displaying reduced susceptibility to overfitting compared to LSTM. Consequently, the GRU encoder with a linear decoder structure generally outperforms the other models.
\begin{table}[ht]
	\centering
	\renewcommand\arraystretch{1.2}
	\fontsize{7}{8}\selectfont
	\caption{Evaluation of Manipulation model with different loss design}
	\begin{tabular}{p{3.5cm}<{\centering}ccccc}
		\toprule
		\multicolumn{1}{c}{\multirow{2}{*}{\parbox{1.1cm}{\centering imitation error}}}&
		\multicolumn{2}{c}{Easy Scenes}&
		\multicolumn{2}{c}{Hard Scenes}\cr
		\cmidrule(lr){2-3} \cmidrule(lr){4-5}
		&$dis_e$&$a_e$&$dis_e$&$a_e$\cr
		\midrule
		Transformer encoder+linear decoder &0.3158&0.3533&0.3556&0.3935\cr
		RNN encoder+linear decoder &0.30535&0.3234&0.3722&0.3681\cr
		LSTM encoder+linear decoder&0.2968&0.3223&0.3521&0.3741\cr
		GRU encoder+linear decoder&0.2748&0.3234&0.3339&0.3482\cr
		\bottomrule
	\end{tabular}
	\label{table:action_imitation}
\end{table}

\subsection{Physical Experimental Demonstration}

\begin{figure}[ht]
        \vspace{0.2cm}
	\centering
	\includegraphics[width=0.6\linewidth]{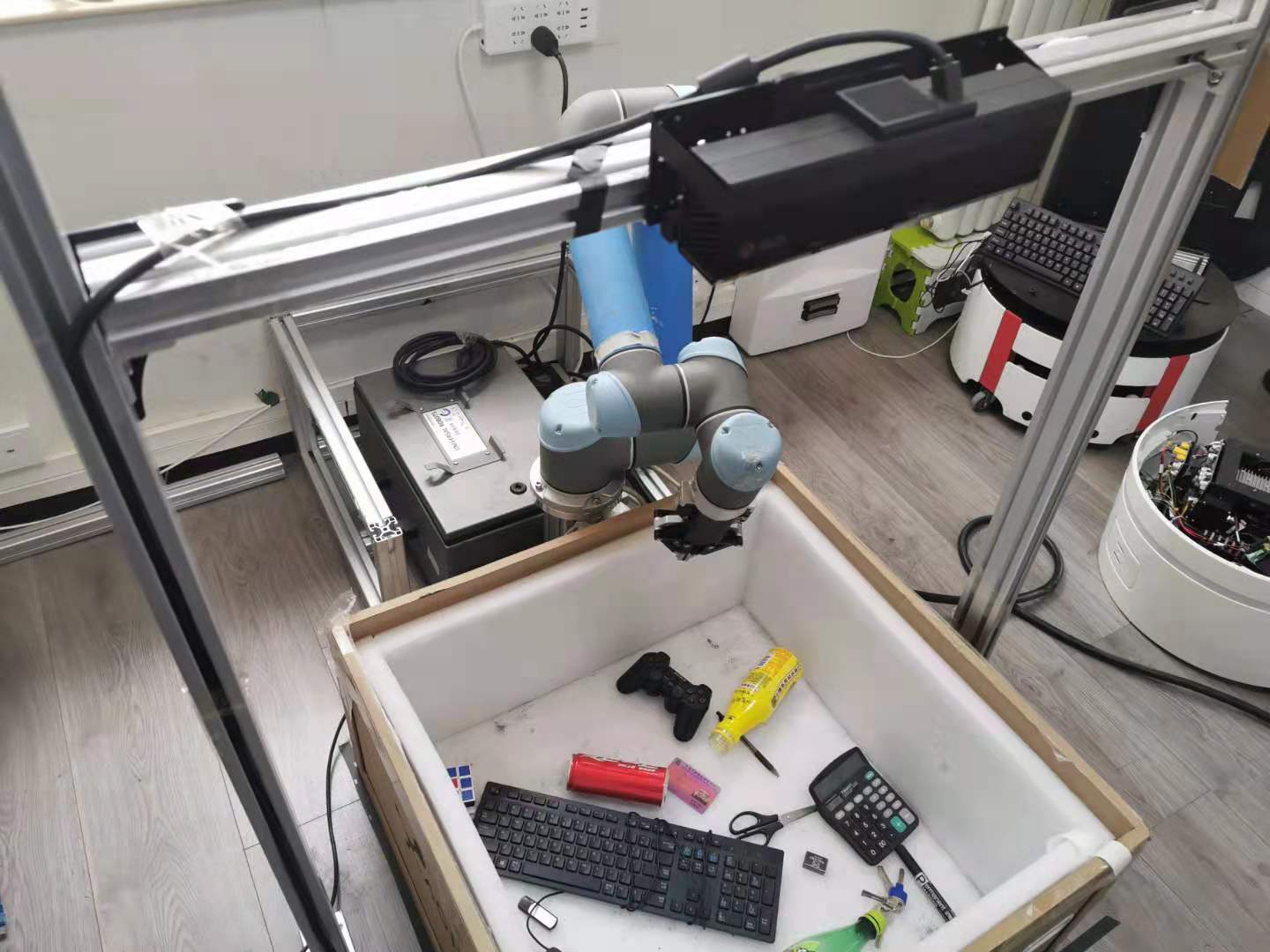}
	\caption{Our physical experimental setup consists of a UR manipulator, a Kinect camera and a bin that was set as the test sample}
	\label{fig:scene-physical}
\end{figure}

We test our proposed method in a physical environment with the setup shown in Fig.~\ref{fig:scene-physical}. We choose a UR manipulator to completer several selected MQA tasks from our test set.  In each task, we arrange a similar layout of objects with corresponding test samples in the simulator.

\begin{figure}[ht]
	\centering
	\includegraphics[width=0.8\linewidth]{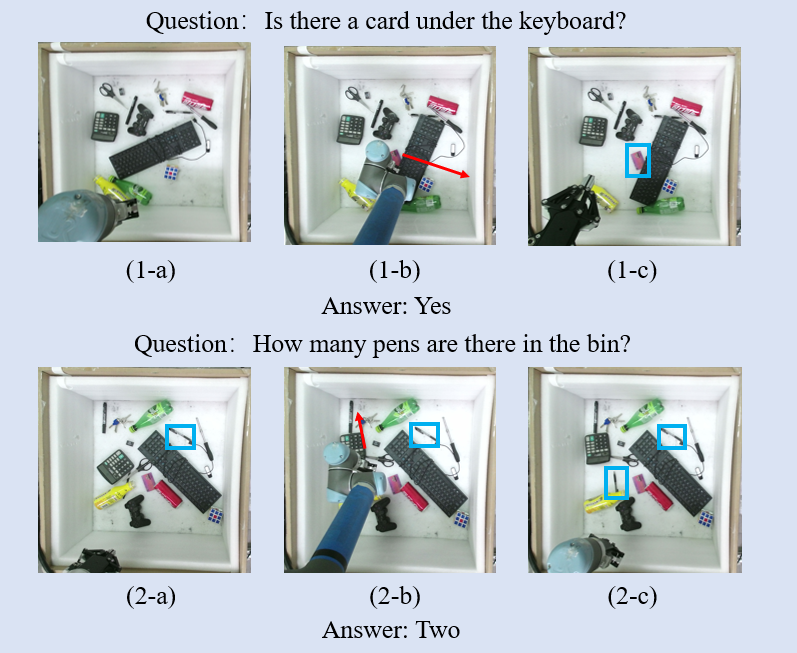}
	\caption{Successful cases of physical experiments:Exist(1-a,1-b,1-c) and Count(2-a,2-b,2-c) Tasks}
	\label{fig:exp-physical}
\end{figure}

\begin{figure}[ht]
	\centering
	\includegraphics[width=0.8\linewidth]{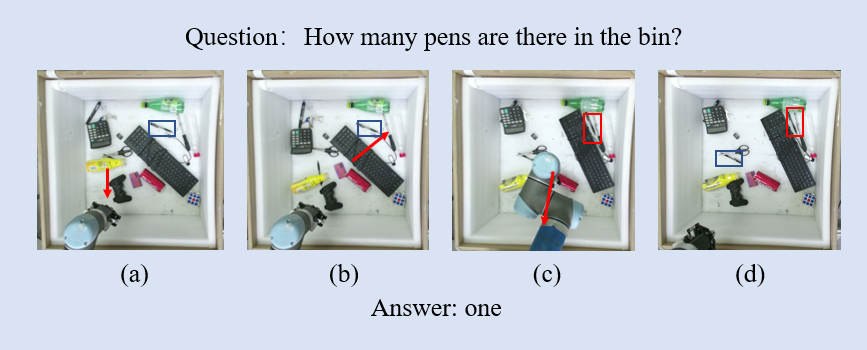}
	\caption{Failure case in which model failed to recognize a pen after relocation}
    \label{fig:fail}
\end{figure}

 Some of selected tasks are successful while others are not. Here we separately present two successful cases in Fig.\ref{fig:exp-physical} and meanwhile one failure case in Fig.\ref{fig:fail}. In successful cases, the manipulator directly moves to the specified object and performs manipulation to answer EXISTENCE questions. In COUNTING questions, the manipulator selects various objects for manipulation until it verifies the presence of all instances in the target category. In a failed scenario, a pen initially sits between the green bottle and the keyboard. The manipulator executes multiple manipulations in an attempt to locate the second pen beneath the scissors and relocate the first pen to the upper right corner of the bin, resulting in a failure recognition. This failure is presumed to arise from a misalignment when relocating the key object with its former location. In densely stacked configurations, objects are in close proximity, potentially causing failures in our criterion, as described in Section \ref{section: model}. 
 

\section{CONCLUSION}
\label{section: conclusion}
In this paper, we proposed a general framework employing a dynamic scene graph for embodied exploration tasks in cluttered scenarios. To test the effectiveness of our proposed method, we choose the MQA task as a test benchmark where the embodied robot implements the question-answering task by manipulating stacked objects in a bin. We designed a manipulation module based on imitation learning and a VQA model based on the dynamic scene graph to solve this task. Extensive experiments on the dataset containing three types of MQA questions in bin scenarios have demonstrated that active exploration with scene graphs is highly effective for answering the question in some cluttered scenarios. The experimental results also prove the rationality of our framework design.










\bibliographystyle{ieeetr}
\bibliography{root.bbl}

\end{document}